\documentclass[lettersize,journal]{IEEEtran}
\usepackage{amsmath,amsfonts}
\usepackage{algorithmic}
\usepackage{algorithm}
\usepackage{array}
\usepackage[caption=false,font=normalsize,labelfont=sf,textfont=sf]{subfig}
\usepackage{textcomp}
\usepackage{stfloats}
\usepackage{url}
\usepackage{verbatim}
\usepackage{graphicx}
\usepackage{cite}
\usepackage{multirow}
\usepackage{amssymb} 
\begin{document}

\title{Adaptive Prompt Learning with SAM for Few-shot Scanning Probe Micoscope Image Segmentation}

\author{Yao Shen, Ziwei Wei, Chunmeng Liu, Shuming Wei, Qi Zhao, Kaiyang Zeng, and Guangyao Li
\thanks{Yao Shen, Chunmeng Liu and Guangyao Li are with College of Electronic and Information Engineering, Tongji University, Shanghai 201804, China.} 
\thanks{Ziwei Wei, Shuming Wei and Kaiyang Zeng are with the Department of Mechanical Engineering, National University of Singapore, Singapore 117575, Singapore.}
\thanks{Qi Zhao is with the Department of Materials Science and Engineering, National University of Singapore, Singapore 117575, Singapore.}}



\maketitle

\begin{abstract}
The Segment Anything Model (SAM) has demonstrated strong performance in image segmentation of natural scene images. However, its effectiveness diminishes markedly when applied to specific scientific domains, such as Scanning Probe Microscope (SPM) images. This decline in accuracy can be attributed to the distinct data distribution and limited availability of the data inherent in the scientific images. On the other hand, the acquisition of adequate SPM datasets is both time-intensive and laborious as well as skill-dependent. To address these challenges, we propose an Adaptive Prompt Learning with SAM (APL-SAM) framework tailored for few-shot SPM image segmentation. Our approach incorporates two key innovations to enhance SAM: 1) An Adaptive Prompt Learning module leverages few-shot embeddings derived from limited support set to learn adaptively central representatives, serving as visual prompts. This innovation eliminates the need for time-consuming online user interactions for providing prompts, such as exhaustively marking points and bounding boxes slice by slice; 2) A multi-source, multi-level mask decoder specifically designed for few-shot SPM image segmentation is introduced, which can effectively capture the correspondence between the support and query images. To facilitate comprehensive training and evaluation, we introduce a new dataset, SPM-Seg, curated for SPM image segmentation. Extensive experiments on this dataset reveal that the proposed APL-SAM framework significantly outperforms the original SAM, achieving over a 30\% improvement in terms of Dice Similarity Coefficient with only one-shot guidance. Moreover, APL-SAM surpasses state-of-the-art few-shot segmentation methods and even fully supervised approaches in performance. Code and dataset used in this study will be made available upon acceptance.
\end{abstract}


\section{Introduction}
Scanning Probe Microscopy (SPM) techniques, encompassing all the tip-based microscopy methods such as Scanning Tunneling Microscopy (STM) and Atomic Force Microscopy (AFM), offer powerful imaging and characterization techniques that have revolutionized the understanding of material science, condensed matter physic, and device operations at micro to nanoscales\cite{meyer2004scanning,rao2011three,dufrene2017imaging}. SPM enables the construction of detailed three-dimensional images of scanned surfaces with nanoscale resolution, capturing various characteristics of the measured materials, such as topography and various functional properties by providing multi-channel data and images. However, analyzing SPM images is always a hot and challenging topic as the images contain more information rather than surface topography, and many times the information is coupled and difficult to be separated. In recent years, deep learning-based analysis are applied to the SPM images\cite{farley2020improving,chang2023deep,lu2023image,krull2020artificial,sotres2021enabling}. Among which, image segmentation in this context involves the application of computational methods to differentiate structural or material features, such as isolating specimens from the background\cite{farley2020improving} for morphology analysis\cite{chang2023deep}, or detecting anomalies at the nanoscale\cite{lu2023image}, in which provides the foundation for the image analysis processes.

Recent advancements in deep learning have enabled the development of automated segmentation tools that can achieve near-human accuracy with exceptional speed. These tools generally rely on extensive training with large amounts of pixel-wise annotated data to reach optimal performance. However, the acquisition and annotation of SPM images are highly complex processes that depend on the materials conditions, set-up of the machine parameters, as well as the expertise and knowledge of skilled professionals. Consequently, exploring the application of few-shot learning\cite{shaban2017one,tian2020prior,li2021adaptive} in SPM image analysis holds substantial practical value but its application in SPM image segmentation remains in its infancy. Existing segmentation algorithms are primarily based on conventional convolutional neural networks under fully-supervision\cite{farley2020improving,chang2023deep,lu2023image}. Moreover, current few-shot learning frameworks for sparse data domains predominantly pretrain their models using data from one single domain, often with restricted datasets, making it difficult to transfer to SPM data\cite{tang2021recurrent,dawoud2021few,akers2021rapid}. Therefore, under few-shot learning paradigm, developing an effective image segmentation model with strong cross-domain feature extraction capabilities is crucial for advancing SPM image segmentation.

Recently, the Segment Anything Model (SAM)\cite{kirillov2023segment}, trained with over 1.1 billon masks from diverse natural image domains, gained significant attention upon its introduction. SAM has demonstrated powerful feature extraction and strong generalization capabilities, enabling the segmentation of any target object within a given image. However, despite its excellent zero-shot transferability, SAM struggles to address challenges associated with specific downstream domains, including medical images, electron microscopy images and SPM images\cite{he2023accuracy,wang2024samda,zhao2023mfs}. This limitation arises because these scientific domains fall outside its training distribution. To enhance SAM's adaptability to these specialized scenarios, previous studies have introduced various strategies. Some have upgraded the architecture of SAM\cite{tianrun2023sam,hu2023skinsam}, either by integrating lightweight Adapters or by refining solely the mask decoder, while others have attempted to combine SAM with additional feature extractors to enable faster domain adaptation\cite{wang2024samda}. However, these methods are not specifically tailored to SPM images, highlighting the urgent need for a SAM-based image segmentation algorithm designed specifically for SPM data, and this has motivated our study presented in this paper.
\begin{figure*}[!t]
\centering
\includegraphics[width=\textwidth]{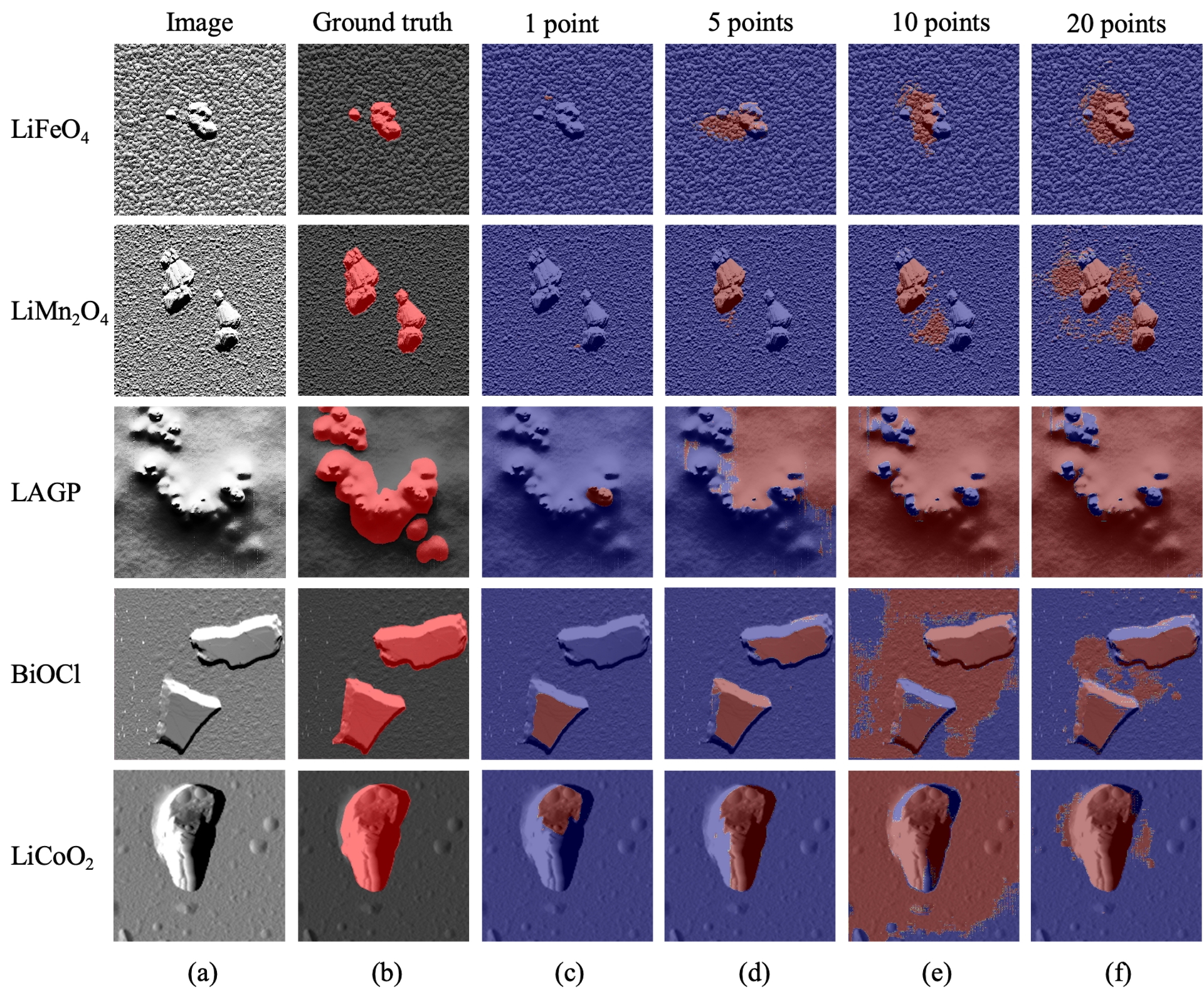}
\caption{The segmentation results produced by the original Segment Anything Model (SAM) on SPM-Seg dataset. (a) represents input images; (b) is corresponding ground truths; and (c)-(f) illustrate the prediction results of the original SAM when using 1-point prompt, 5-point prompts, 10-point prompts and 20-point prompts, respectively (here, $LAGP$ is $Li_{1.5}Al_{10.5}Ge_{1.5}(PO_{4})_{3}$ solid electrolyte, $LiFeO_{4}$, $LiCoO_{2}$ and $LiMn_{2}O_{4}$ are electrode materials for Li-ion battery, and $BiOCl$ is a crystal powder). Here, the scan size for sample $LiFeO_{4}$, sample $LAGP$ and sample $LiCoO_{2}$ is $5\mu m \times5\mu m$, and the scan size for sample $LiMn_{2}O_{4}$ and sample $BiOCl$ are $3\mu m \times3\mu m$ and $2\mu m \times2\mu m$, respectively. }
\label{fig_1}
\end{figure*}
When the original SAM is directly applied to the SPM images of various materials, the predictions are suboptimal (Fig. 1). The model is easily disturbed by noise, fuzzy boundaries, minor foreground-background differences and ‘shadows’, resulting in misclassifications and false negatives. Here, we define ‘shadows’ as the phenomenon similar to light shadow caused by differential signals in the data from amplitude channel during the SPM measurements. To eliminate the influences of these factors and to achieve accurate and automatic segmentation without vast training data, we propose the Adaptive Prompt Learning with SAM framework for SPM image segmentation, named as APL-SAM in this work. Our new framework adopts the classical two-branch few-shot segmentation structure\cite{shaban2017one}, utilizing the meta-learning training paradigm\cite{antoniou2018train}, and it comprises two core components. The first one is a revamped prompt-encoder that employs the Adaptive Prompt Learning module to leverage few-shot embeddings derived from limited support set, adaptively forming visual prompts based on the target scales and shapes. The second component is an enriched mask decoder, which integrates information from both the support and query images and conducts predictions of multiple levels. Additionally, we incorporate Adapters to fine-tune SAM's image encoder, following methods proposed by Houlsby et al.\cite{houlsby2019parameter}, to enable SAM to effectively learn features specific to SPM data. Moreover, we have constructed a SPM dataset, named as SPM-Seg, consisting of 195 SPM images with pixel-wise labeled annotations, to train the proposed framework. The core contributions of this study can be summarized as follows:
\begin{enumerate}
\item{We develop an enhanced SAM-based few-shot segmentation framework, APL-SAM. This new framework is designed to achieve accurate, automatic and robust segmentation of SPM images with minimal training effort, facilitating morphology analysis at the micro-nano scales.}
\item{We introduce a modified prompt encoder based on the Adaptive Prompt Learning module, which forms visual prompts adaptive to different target scales and shapes, and integrate a multi-source, multi-level mask decoder for more precise predictions.}
\item{We construct a SPM dataset, SPM-Seg, comprising 195 SPM images and the corresponding annotations, which is open-source to advance further data-based SPM image analysis.}
\item{Our method outperforms SAM by over 30\% in terms of Dice Similarity Coefficient and surpasses both other state-of-the-art few-shot segmentation methods and the fully-supervised segmentation frameworks.}
\end{enumerate}

\section{Related Work}
\subsection{Scanning Probe Microscope Image Segmentation}
Semantic segmentation in Scanning Probe Microscope images enables the extractions of regions of interest from the complex backgrounds, which is of great significance for further quantitative analysis of the images at various levels\cite{bui2020segmentation}. Although the manual segmentation methods can be applied to SPM images, they are inefficient and susceptible to human bias\cite{barrera2008automated}. To overcome these limitations, a range of automated segmentation methods have been developed\cite{bui2020segmentation,liu2022feasible,raila2022ai,klapetek2011atomic,venkataraman2006automated}, primarily categorized into threshold-based methods\cite{liu2022feasible,raila2022ai}, which classify pixels into foreground and background by setting specific thresholds; and watershed algorithms\cite{klapetek2011atomic,venkataraman2006automated}, which invert the morphology height to delineate regions. When processing raw SPM data, however, these methods are often inadequate due to artifacts, noise, and distortions in the images caused by various factors such as vertical drift, hysteresis, cross-coupling, and others\cite{chang2023deep}. With the advancement of artificial intelligence, image segmentation algorithms driven by deep learning have led to substantial progress in the SPM image analysis\cite{farley2020improving,chang2023deep,sotres2021enabling,gordon2020automated,alldritt2020automated}. Fully convolutional networks have been adapted for SPM domains to enhance accuracy and robustness\cite{farley2020improving,long2015fully}, while U-net-like architectures have been employed to extract enriched features from raw SPM images\cite{chang2023deep,ronneberger2015u,zhang2021lcu}. Despite these advancements, the performance of such methods relies heavily on the data volume and label density. Developing approaches that minimize this reliance on extensive data holds significant practical importance. In this study, we integrate the few-shot learning mechanism into the SPM image segmentation enabling high accuracy with only one-shot guidance.
\subsection{Foundation Models}
Foundation models are extensively pre-trained models built using self-supervised learning techniques and trained on large-scale datasets from various domains. This facilitates swift adaptation to specific down-stream tasks through fine-tuning or in-context learning. In the field of Natural Language Processing (NLP), foundation models have undergone several advancements and have gradually reached maturity\cite{devlin2018bert,brown2020language,achiam2023gpt}. In the range of computer vision, foundation models are also redefining traditional research paradigms and sparking new revolutions, resulting in models such as CLIP\cite{radford2021learning}, ALIGN\cite{jia2021scaling} and Blip-2\cite{li2023blip}. The Segment Anything Model (SAM) is the pioneering foundation model for visual segmentation\cite{kirillov2023segment}, having been pre-trained on 11 million images with more than 1.1 billion masks. This interactive model enables users to specify the targets by providing various geometric prompts, such as points, boxes, masks and texts, yielding the corresponding accurate segmentation results in any images. Although SAM demonstrates its powerful zero-shot transferability across conventional image domains, it performs sub-optimally in multiple downstream tasks, including the segmentation of SPM images.
\subsection{Enhanced SAM for Downstream Tasks}
Based on the limitations of SAM, some research concentrates on enhancing its generalization performance and generating higher-quality predictions\cite{li2023semantic,wang2024sam,ke2024segment}, while others aim to adapt SAM to specific downstream domains\cite{chen2023sam,song2024simada,ma2024segment,wu2023medical}. For example, Med-SAM\cite{ma2024segment} refines SAM’s mask decoder by training it on a carefully curated medical image dataset with over 200,000 masks across 11 different modalities. Med SAM Adaptor\cite{wu2023medical} improves SAM by incorporating a series of adaptor neural network modules into the original SAM architecture. These fine-tuning approaches require considerable training effort, including substantial dataset preparation and computational resources, which severely degrades SAM’s zero-shot capabilities. In contrast, the method proposed in this paper utilizes a few-shot learning mechanism, significantly minimizing the training effort needed and offering a practical solution for adapting SAM to SPM image segmentation.

\section{Methodology}
\subsection{Problem definition}
We address the challenge of few-shot SPM image segmentation, which involves segmenting previously unseen classes in query SPM images using only a small number of support SPM images. In this regard, we adopt a meta-learning approach that constructs a meta-learner $M$ to handle a range of few-shot SPM image semantic segmentation tasks $T=\{T\}$ drawn from an underlying task distribution. Following that proposed in \cite{leng2024self}, we implement a MAML++\cite{antoniou2018train} based meta-learner to identify optimal initialization parameters, facilitating rapid adaptation to various materials. 
\begin{figure}[!t]
\centering
\includegraphics[width=3.5in]{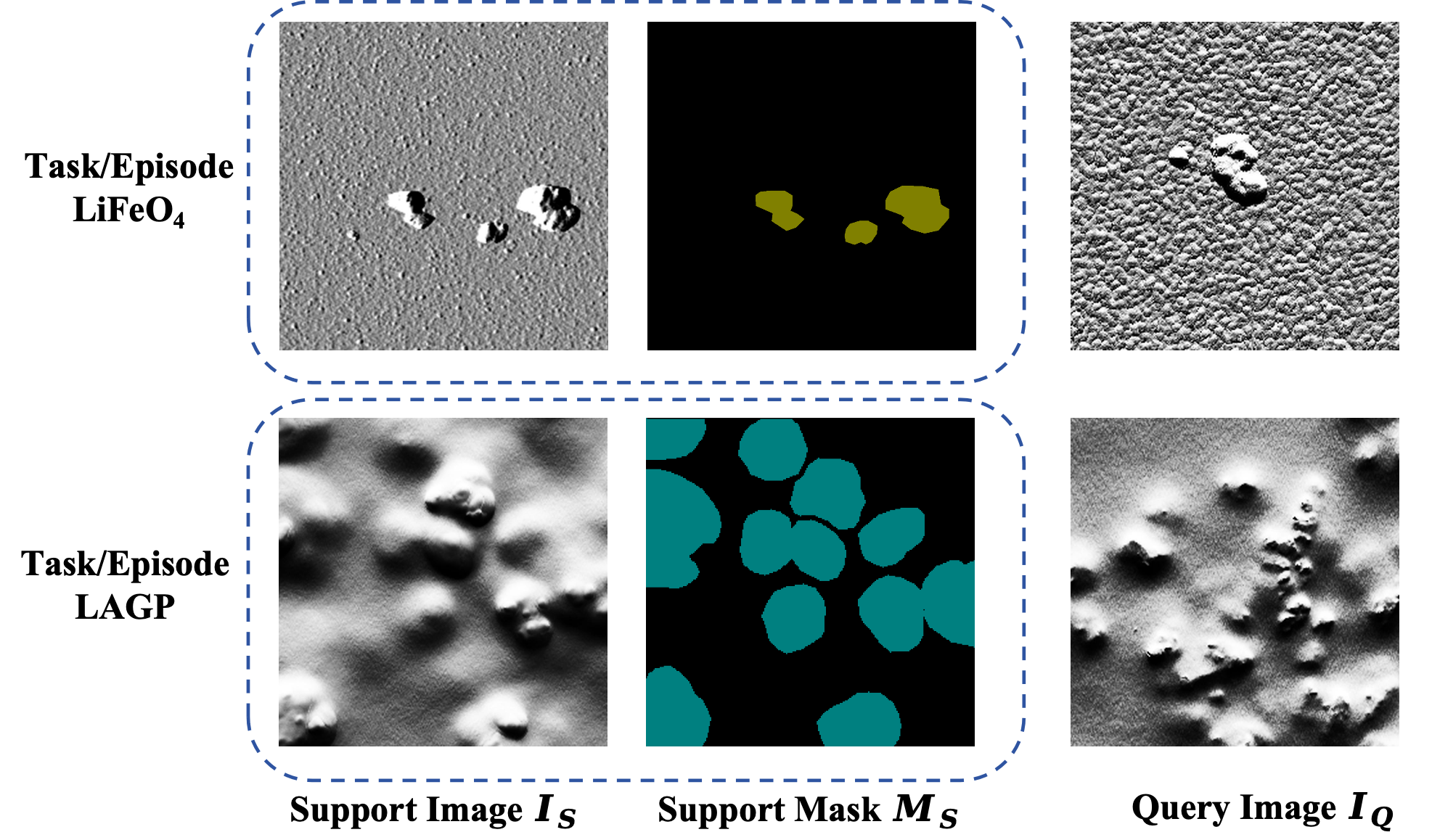}
\caption{Examples of the one-way one-shot segmentation tasks. For each material class, an episode(task) consists of a support set $S$ (denoted by the dashed box) and a query image $I_{Q}$. Each support set includes one support image $I_{S}$ along with its corresponding mask $M_{S}$ (one-shot). In each episode, only a single material class is targeted for segmentation (one-way). Here, the scan size for both $LiFeO_{4}$ and $LAGP$ is $5\mu m \times5\mu m$.}
\label{fig_2}
\end{figure}

In this study, we formulate the few-shot segmentation problem into a one-way one-shot segmentation task, where only one labeled support image provides the supervision to segment the target class (Fig. 2). Specifically, each SPM image segmentation task $T$ is composed of a support set $S= \{I_{S},M_{S}\}$, where $I_{S}$  is the support image and $M_{S}$ is the corresponding mask, and a query image $I_{Q}$ of the same class c. The model is meta-trained on classes $C_{train}$ and meta-tested on previously unseen classes $C_{test}$ ($C_{train} \cap C_{test}= \varnothing $). The ground truth mask $M_{Q}$ of the query image is not accessible to the model and is solely used for evaluating predictions on the query image in each episode\cite{vinyals2016matching}.

\subsection{Overview of APL-SAM Architecture}

\begin{figure}[!t]
\centering
\includegraphics[width=3.4in]{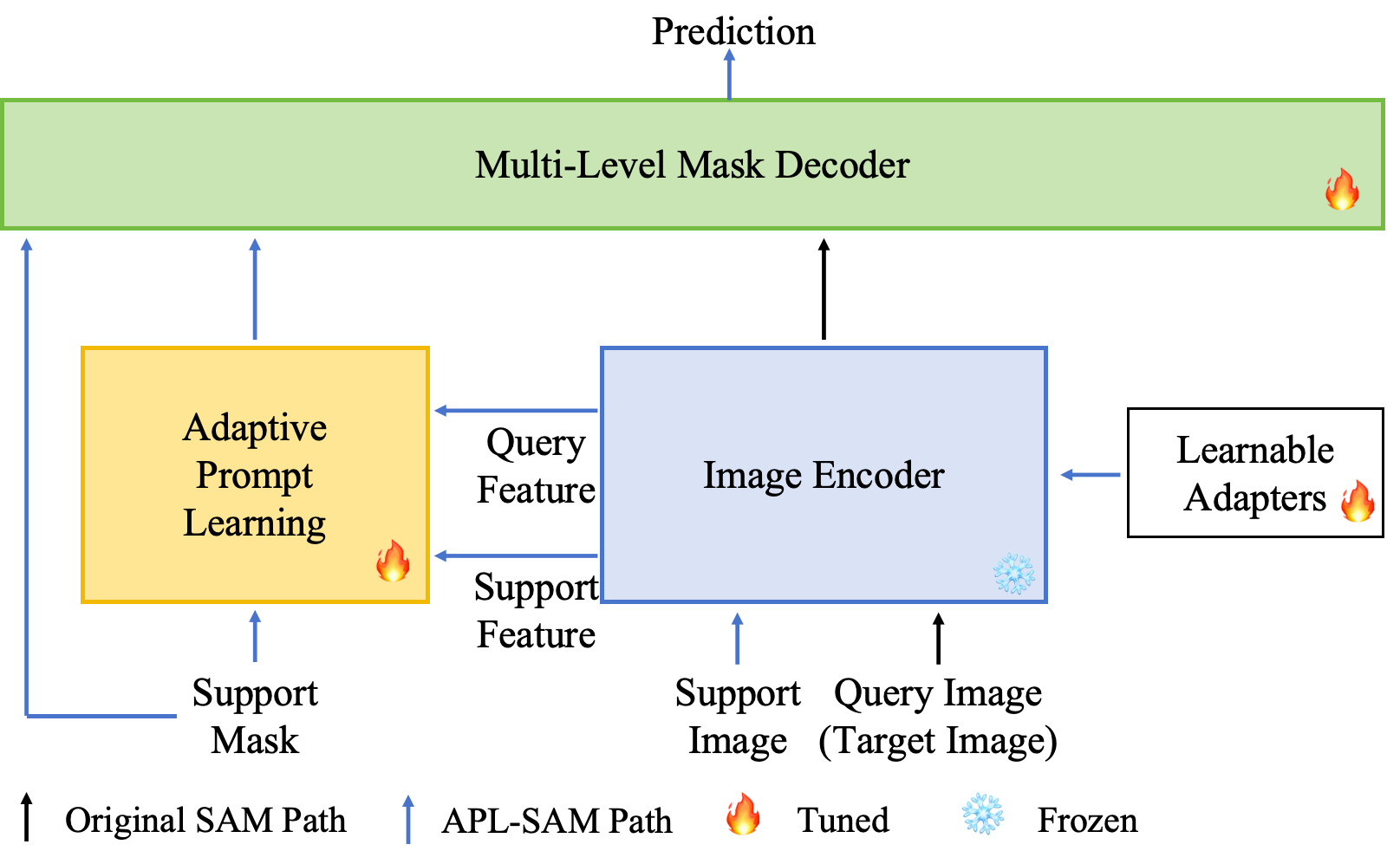}
\caption{Overview of the APL-SAM architecture proposed in this study. The image encoder of SAM remains frozen, while a series of lightweight adapters are introduced to adapt to the SPM images. APL-based prompt encoder and the multi-level mask decoder are tuned during training.}
\label{fig_3}
\end{figure}

The original SAM\cite{kirillov2023segment} employs three core modules for processing each input image: a heavyweight image encoder (i.e., Vision Transformer (ViT)\cite{dosovitskiy2020image} as the backbone) that extracts image embeddings, a prompt encoder that generates prompt embeddings from the geometric prompts (i.e., points, a bounding-box, or a coarse mask), and a lightweight mask decoder that integrates image embeddings and prompt embeddings to produce the segmentation predictions. Fig. 3 illustrates the architecture of the few-shot segmentation framework proposed in this study. We retain the original SAM image encoder as fixed and introduce a learnable Adapter layer\cite{houlsby2019parameter} to each transformer layer for optimized training efficiency. Furthermore, we divide all the transformer layers of the image encoder into four parts, each comprising three layers, with their outputs sequentially processed for subsequent steps. Following processing by the image encoder, we perform the adaptive prompt learning in prompt encoder part, utilizing superpixels in the masked support feature and the cross-attention to extract visual prompts. The multi-level mask decoder then integrates the four different levels of output from the image encoder along with the support prompt to generate the final prediction.
\subsection{Image Encoder with Adapters}
The image encoder of SAM, based on a ViT\cite{dosovitskiy2020image} pretrained on Masked Autoencoders (MAE)\cite{he2022masked}, possesses strong feature extraction capabilities but is computationally heavy. In this study, we utilize the ViT-B variant, which comprises 12 transformer layers, to generate the final image embeddings. To facilitate efficient learning with faster updates and alleviate the strain on GPU memory, we keep the image encoder frozen and insert a lightweight sub-network, referred to as an Adapter, into each transformer layer. Given an input image $I$ of any size, we first resize it to a resolution of $1024\times1024$ and then partition it into patches to generate patch embeddings $E^{0}$. The Adapters, designed to learn the task-specific features and capture low-level structural information, adjust the input of each transformer layer. Every Adapter consists of a linear down-projection $MLP_{down}^{i}$, a nonlinear activation function Act, a linear up-projection $MLP_{up}$, and a residual connection. The $MLP_{up}$ is shared across all Adapters and is crafted to match the dimension of transformer features. To formulate the input for the Adapters, we extract the high-frequency components $I_{hfc}$ of the image $I$ and then reverse it to the space domain using the Fast Fourier Transform and its inverse. Subsequently, convolutional layers and linear layers project $I_{hfc}$ as well as the original image embeddings $E^{0}$ into c-dimensional features independently, yielding $F_{hfc}$ and $F_{pe}$, respectively. The result of element-wise addition of $F_{hfc}$ and $F_{pe}$ is then forwarded to the $i$-th Adapter. We set the output of the $i$-th Adapter as $F_{A}^{i}$ and hence:
\begin{equation}
\label{eq1}
F_{A}^{i} = MLP_{up}(Act(MLP_{down}^{i}(F_{hfc} + F_{pe}))).
\end{equation}
The input to the $i$-th transformer layer $F^{i}$ is formulated as:
\begin{equation}
\label{eq2}
F^{i} = E^{i-1} + F_{A}^{i}
\end{equation}
where $E^{i-1}$ is the output from the previous transformer layer. In addition, we divide the 12 layers of the image encoder into four sub-blocks, with outputs generated after every 3 layers. In this way, multi-level information of the query image can be extracted.

\subsection{Adaptive Prompt Learning}
\begin{figure}[!t]
\centering
\includegraphics[width=3.4in]{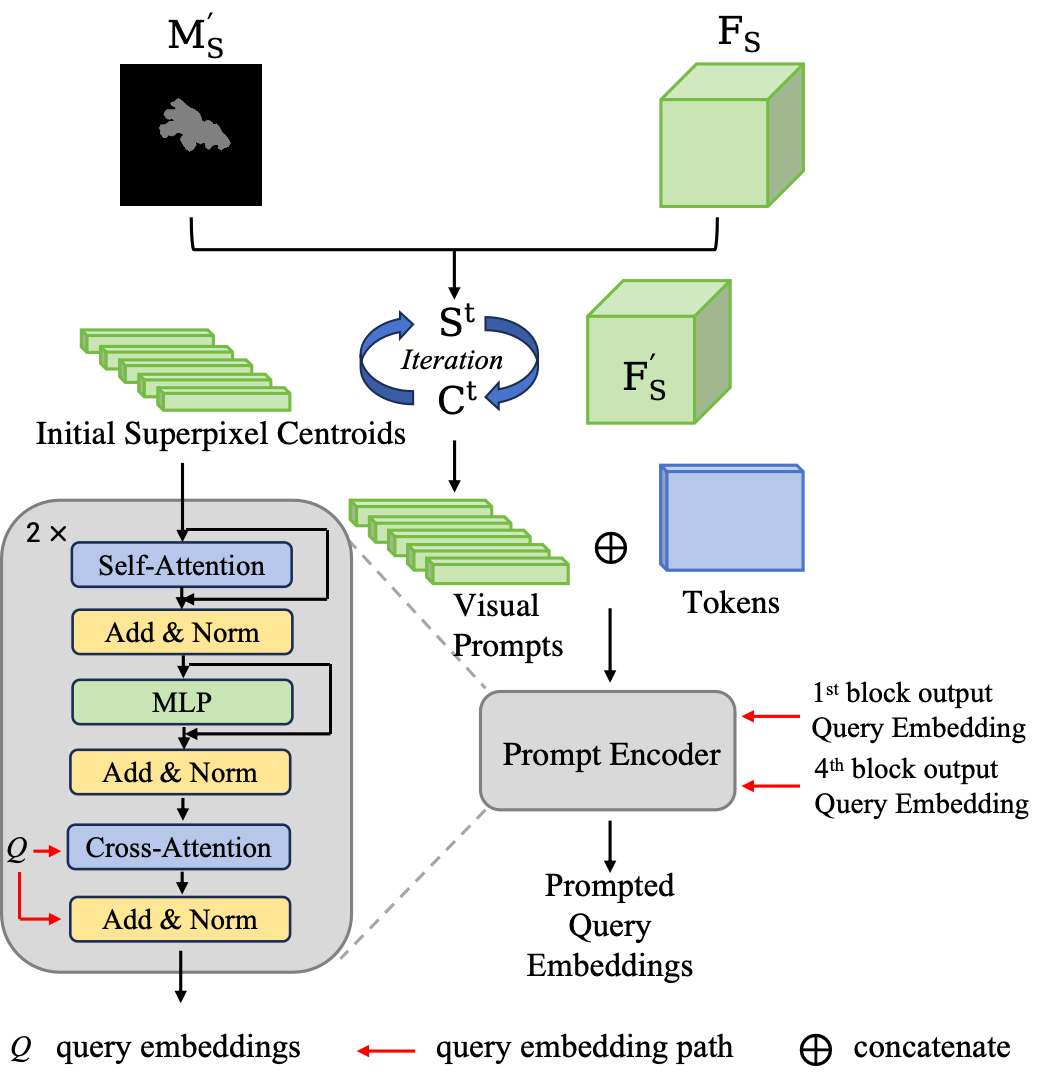}
\caption{Architecture of the Adaptive Prompt Learning based prompt encoder proposed in this study.}
\label{fig_4}
\end{figure}
Fig. 4 shows the entire process of the Adaptive Prompt Learning-based prompt encoder. We replace the user-provided interactive prompts with an Adaptive Prompt Learning module that extracts visual prompts from the few-shot support set. Drawing inspiration from \cite{li2021adaptive}, we employ a superpixel-guided clustering method to elicit the most representative prototypes, serving as the visual prompts. A superpixel is defined as a set of pixels that share similar characteristics such as color, texture or category, and it is often regarded as a basic unit for the image segmentation task. We derive the centroids of superpixels in the high dimensional embedding space as information aggregation points, which functions as the aforementioned visual prompts. To align with the query embeddings in the same embedding space, we also utilize the image encoder is to extract support embedding from the support image. 

Given the support embedding $F_{S} \in R^{c\times h\times w}$, we first downsample the corresponding support mask $M_{S}$  to $M_{S}^{'} \in R^{h\times w}$, matching the resolution of $F_{S}$. This downsampled mask $M_{S}^{'}$ is then used to filter out the background information in the support embedding, compressing $F_{S} \in R^{c\times h\times w}$ to $F_{S}^{'} \in R^{c\times N_{m}}$, where $N_{m}$ is the number of pixels within the foreground region. To uniformly set initial centroids in the masked foreground region, we follow the MaskSLIC\cite{irving2016maskslic} to determine the coordinates of initial centroids $C^{0}$ by using $M_{S}^{'}$. In this way, the initial centroids $C^{0} \in R^{(c+2)\times N_{c}}$ can be sampled from the support embedding as well as their coordinates, where $N_{c}$ denotes the number of the centroids.

Subsequently, the final superpixel centroids is computed through an iterative process. We define the arithmetic square root of the feature distance and spatial distance as the integrated distance $D$\cite{achanta2012slic}:
\begin{equation}
\label{eq3}
D = \sqrt{(d_{f})^{2} + (d_{s}/\omega)^{2}}
\end{equation}
where $d_{f}$ represents the Euclidean distance for features, $d_{s}$ is the Euclidean distance for coordinate values, and $\omega$ serves as a weighting factor. For each iteration $t$, we first calculate the corresponding map $S^{t}$ between each pixel $p$ and all superpixel centroids following the integrated distance:
\begin{equation}
\label{eq4}
S_{p,i}^{t} = e^{-D(F_{p}^{'},C_{i}^{t-1})}=e^{-\parallel F_{p}^{'}-C_{i}^{t-1} \parallel^{2}}.
\end{equation}
Then, new superpixel centroids replace the old ones by the weighted sum of masked features:
\begin{equation}
\label{eq5}
C_{i}^{t} = \frac{1}{\sum_{p}S_{p,i}^{t}}\sum_{p=1}^{N_{m}}S_{p,i}^{t}F_{p}^{'}.
\end{equation}
After completing all the iterations, the final superpixel-based centroids $C \in R^{c\times N_{c}}$, representing the visual prompts, are obtained. During training, the number of iterations is set to 10.

We initialize $N_{t}$ tokens $T\in R^{c\times N_{t}}$ to query the knowledge embedded in the visual prompts. These tokens are concatenated with the visual prompts and processed through self-attention to absorb the support knowledge, which is then transferred to the query embeddings via cross-attention. Notably, we select query embeddings from both the first and final block outputs of the image encoder. 

The Adaptability is achieved with respect to the target scale. Given the nature of the SPM images, which reflect the surface morphology of the materials in this study, some foreground targets are relatively small within the scanning range, such as LiCoO2, allowing a single centroid to encapsulate all relevant information. However, for larger target objects with extensive distribution, such as LAGP, multiple centroids are required to adequately capture their characteristics (Fig. 1). To ensure adaptability to the scale of the target object, we make the number of superpixel centroids flexible, defined by:
\begin{equation}
\label{eq6}
N_{c} = \min(\lfloor \frac{N_{m}}{A_{sp}} \rfloor, N_{max}),
\end{equation}
where $N_{m}$ represents the total number of pixels in the masked region, and $A_{sp}$  denotes the average area allocated to each initial superpixel centroid, with an empirical value of 100. $N_{max}$ is a hyperparameter which is set to avoid the computational burden. When the foreground object occupies a small portion of the image, $N_{c}=0$ or 1, in this case, the masked average pooling\cite{zhang2020sg} method is employed to extract the only centroid.
\subsection{Multi-level Attention-based Mask Decoder}
\begin{figure*}[!t]
\centering
\includegraphics[width=5in]{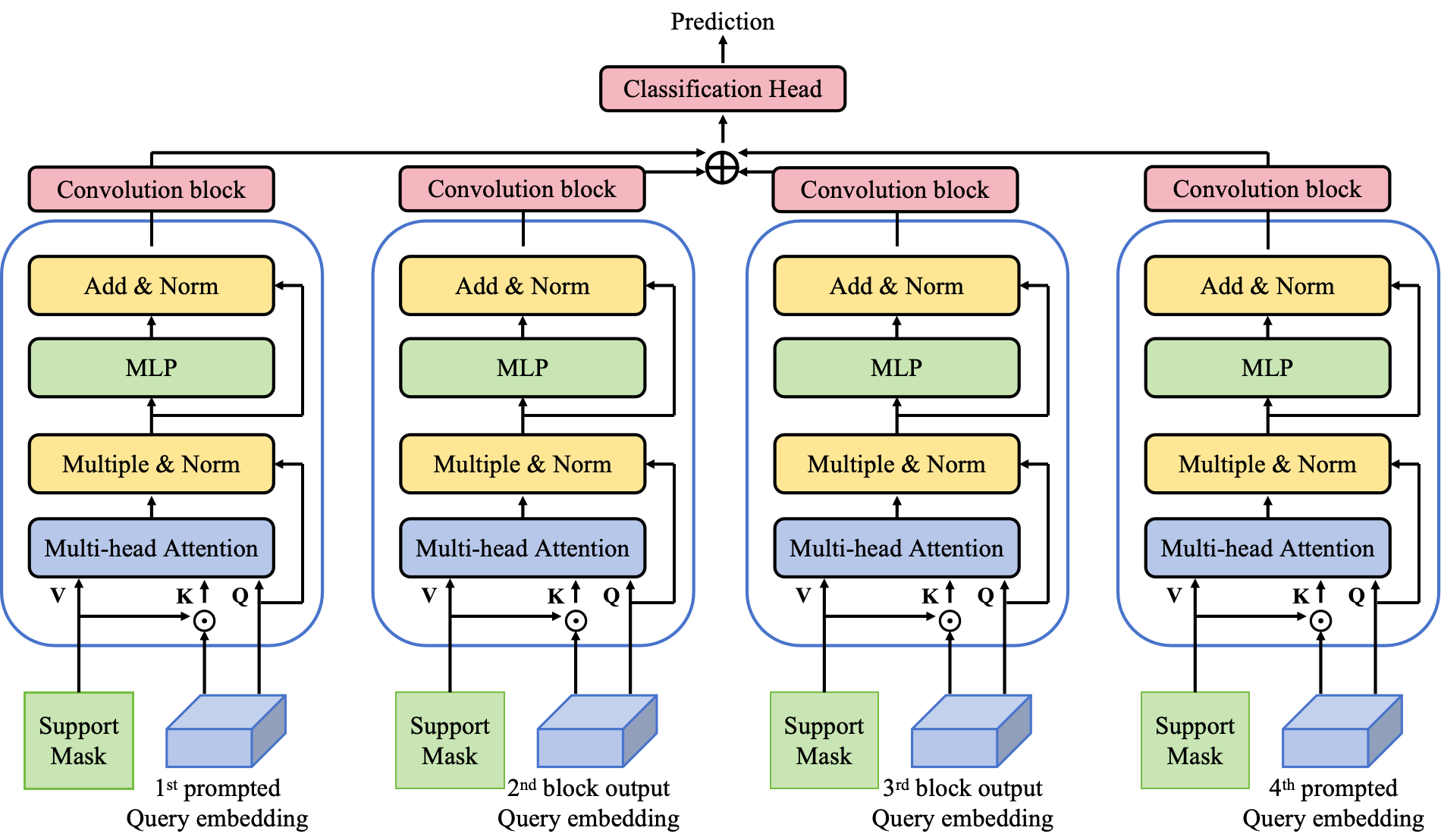}
\caption{Structure of the multi-level mask decoder (MLMD). Four attention-based blocks process the prompted embeddings at different levels in parallel and the final prediction is obtained by integrating all four outputs.}
\label{fig_5}
\end{figure*}
In this study, a multi-source, multi-level mask decoder (MLMD) is applied to process and predict query embeddings output from different levels in parallel and then fuse the results for enhanced segmentation results. Additionally, support masks are incorporated into the decoder network to better activate foreground regions and suppress distractions caused by background, noise and other factors. The complete flowchart of the mask decoder is shown in Fig. 5. We project the element-wise product of the support mask and the multi-level query embeddings as the key input for the cross-attention module. After processing through four attention modules, the outputs are concatenated, and the final prediction is generated by the meticulously designed classification head.

\section{Experiments}
\subsection{Dataset Construction}
Deep learning approaches generally necessitate massive data to adequately train models. However, the limited availability of public datasets in SPM measurements poses a significant challenge to the application of the deep learning algorithms to SPM image analysis. To address this issue, we construct a SPM image dataset, named as \textbf{SPM-Seg}, comprising 195 images along with their corresponding label masks. This dataset encompasses eight material categories, with half representing the materials used for rechargeable battery systems, such as $LiFeO_{4}$ and $LiCoO_2$. This dataset provides both training of the proposed APL-SAM network and also served as a valuable resource for further developing deep learning-based SPM image processing tasks, including image classification and reconstruction. In the SPM experiments, all images in the dataset are obtained from a commercial SPM system (MFP-3D, Asylum Research, Oxford Instruments, CA, USA), using tapping mode with vertical feedback. These images are derived from the amplitude channel output data and converted into 2D images. The self-scanned raw images contain various types of noise and distortion, which have been preserved to enhance the robustness of the model. It is worth noting that the LabelMe software is employed to manually segment each image, generating corresponding labels.

In this study,$LiFeO_{4}$, $LiMn_{2}O_{4}$, $LAGP$, $LiCoO_{2}$ and $BiOCl$ are used as semantic classes. In each experiment, one material is designated as the unseen semantic class for testing, while the remaining materials are utilized for training.
\subsection{Evaluation Metrics and Implementation Details}
In this study, we employ Dice Similarity Coefficient (DSC) and Intersection over Union (IoU) as major evaluation metrics. The DSC measures the overlap between the predicted and ground truth segmentation masks and IoU quantifies the ratio of the intersection to the union of the predicted and ground truth masks. The mathematical formulations for DSC and IoU are as follows:
\begin{equation}
\label{eq7}
DSC = \frac{2\cdot TP}{(TP+TF)+(TP+FN)},
\end{equation}
\begin{equation}
\label{eq8}
IoU = \frac{TP}{(TP+FP+FN)},
\end{equation}
Here, \textit{TP} (True Positive) denotes the area where the predicted mask correctly overlaps with the ground truth; \textit{FN} (False Negative) refers to the part of the target region that is missed by the prediction, and \textit{FP} (False Positive) indicates the portion of the predicted segmentation that incorrectly covers non-target areas. Both metrics range between 0 and 1, where a value closer to 1 indicates a better segmentation performance.

All images are resized to $1024 \times 1024$ to be compatible with the SAM model. We conduct the 1-way 1-shot learning which means each episode contains one support image with label and one query image needs to be predicted. We use ViT-B version\cite{dosovitskiy2020image} of SAM and train our network by minimizing the sum of the balanced cross-entropy loss and IoU loss between the predicted and ground truth masks:
\begin{equation}
\label{eq9}
\begin{aligned}
L &= L_{bce} + L_{iou} \\ 
&=-(\alpha\cdot gt \cdot \log(p) + \beta\cdot(1-gt)\cdot\log(1-p))\\
&+(1-iou).    
\end{aligned}
\end{equation}
Here, $\alpha$ and $\beta$ are adjustment factors, $gt$ denotes the ground-truth and $p$ represents the prediction result. The AdamW optimizer is employed for all experiments and we set the initial learning rate to $5e^{-5}$. The cosine decay schedule is applied to the learning rate and the training is conducted over 50 epochs. Both few-shot segmentation task and fully supervised segmentation task are trained on a single NVIDIA RTX 4090 GPU, using Pytorch.

\subsection{Quantitative Results}
\begin{table*}[!t]
\caption{Quantitative evaluation of the proposed APL-SAM in comparison with representative few-shot segmentation methods, fully-supervised segmentation methods and the original SAM with different number of point prompts, in terms of DSC (unit: \%). \textbf{Best} results are highlighted in \textbf{bold}.\label{tab:table1}}
\centering
\begin{tabular}{|c|c|c|c|c|c|c|}
\hline
Method & $LiFeO_{4}$ & $LiMn_{2}O_{4}$ & $LAGP$ & $BiOCl$ & $LiCoO_{2}$ & mean\\
\hline
SSL-ALPNet\cite{ouyang2020self} & 61.40 & 57.27 & 21.81 & 42.33 & 62.20 & 49.00\\
RPNet\cite{tang2021recurrent} & 70.36 & 66.74 & 27.20 & 46.37 & 70.86 & 56.30\\
\hline
nnUnet\cite{isensee2021nnu} & 91.23 & 87.31 & 58.03 & 85.69 & 87.44 & 81.94\\
\hline
SAM\cite{kirillov2023segment} (with 1 point) & 28.84 & 21.15 & 12.02 & 28.53 & 22.35 & 22.58\\
SAM\cite{kirillov2023segment} (with 20 points) & 75.33 & 53.62 & 42.93 & 80.15 & 24.19 & 55.24\\
\hline
APL-SAM(w/o MLMD) & 94.00 & 91.79 & 72.69 & 88.03 & 92.93 & 87.89\\
APL-SAM & \textbf{95.35} & \textbf{92.60} & \textbf{73.54} & \textbf{90.92} & \textbf{95.24} & \textbf{89.53}\\
\hline
\end{tabular}
\end{table*}

\begin{table*}[!t]
\caption{Quantitative evaluation of the proposed APL-SAM in comparison with representative few-shot segmentation methods, fully-supervised segmentation methods and the original SAM with different number of point prompts, in terms of IoU (unit: \%). \textbf{Best} results are highlighted in \textbf{bold}.\label{tab:table2}}
\centering
\begin{tabular}{|c|c|c|c|c|c|c|}
\hline
Method & $LiFeO_{4}$ & $LiMn_{2}O_{4}$ & $LAGP$ & $BiOCl$ & $LiCoO_{2}$ & mean\\
\hline
nnUnet\cite{isensee2021nnu} & 84.94 & 78.30 & 41.14 & 78.12 & 79.42 & 72.38\\
\hline
SAM\cite{kirillov2023segment} (with 1 point) & 24.83 & 17.05 & 6.68 & 23.45 & 18.38 & 18.08\\
SAM\cite{kirillov2023segment} (with 20 points) & 67.90 & 44.04 & 29.09 & 70.36 & 20.07 & 46.29\\
\hline
APL-SAM(w/o MLMD) & 88.69 & 84.74 & 57.10 & 78.62 & 85.85 & 79.00\\
APL-SAM & \textbf{91.12} & \textbf{86.22} & \textbf{58.16} & \textbf{83.35} & \textbf{90.91} & \textbf{81.95}\\
\hline
\end{tabular}
\end{table*}

Table I and Table II present the quantitative segmentation results using the APL-SAM proposed in this study and comparison with that from other segmentation methods, evaluated using the DSC and IoU metrics, respectively. We first directly apply the original SAM to segment the SPM images, providing randomly selected foreground points as prompts. Even with 20 prompt points, SAM struggled to accurately segment the SPM data, highlighting the necessity of our successful enhancements to adapt SAM for the SPM image domain. The proposed APL-SAM achieves enhancements of 34.29\% and 35.66\% in terms of DSC and IoU, respectively. SSL-ALPNet and RPNet are state-of-the-art few-shot medical image segmentation frameworks. The results illustrate that APL-SAM significantly outperforms SAM and these conventional few-shot methods, achieving over a 30\% improvement in terms of both DSC and IoU. This confirms that, under one-shot support guidance, leveraging the robust feature extraction and exceptional generalization capabilities of SAM leads to superior segmentation of previously unseen classes. nnUNet is one of the state-of-the-art fully supervised segmentation networks which is trained on the entire training data and evaluated on the same segmentation task. Remarkably, our APL-SAM outperforms nnUNet by 7.59\% in terms of DSC and 9.57\% in terms of IoU. These experiments further underscore the superiority of APL-SAM, demonstrating its ability to effectively adapt SAM to SPM images with minimal annotated data requirements.
\subsection{Qualitative Results}
\begin{figure*}[!t]
\centering
\includegraphics[width=\textwidth]{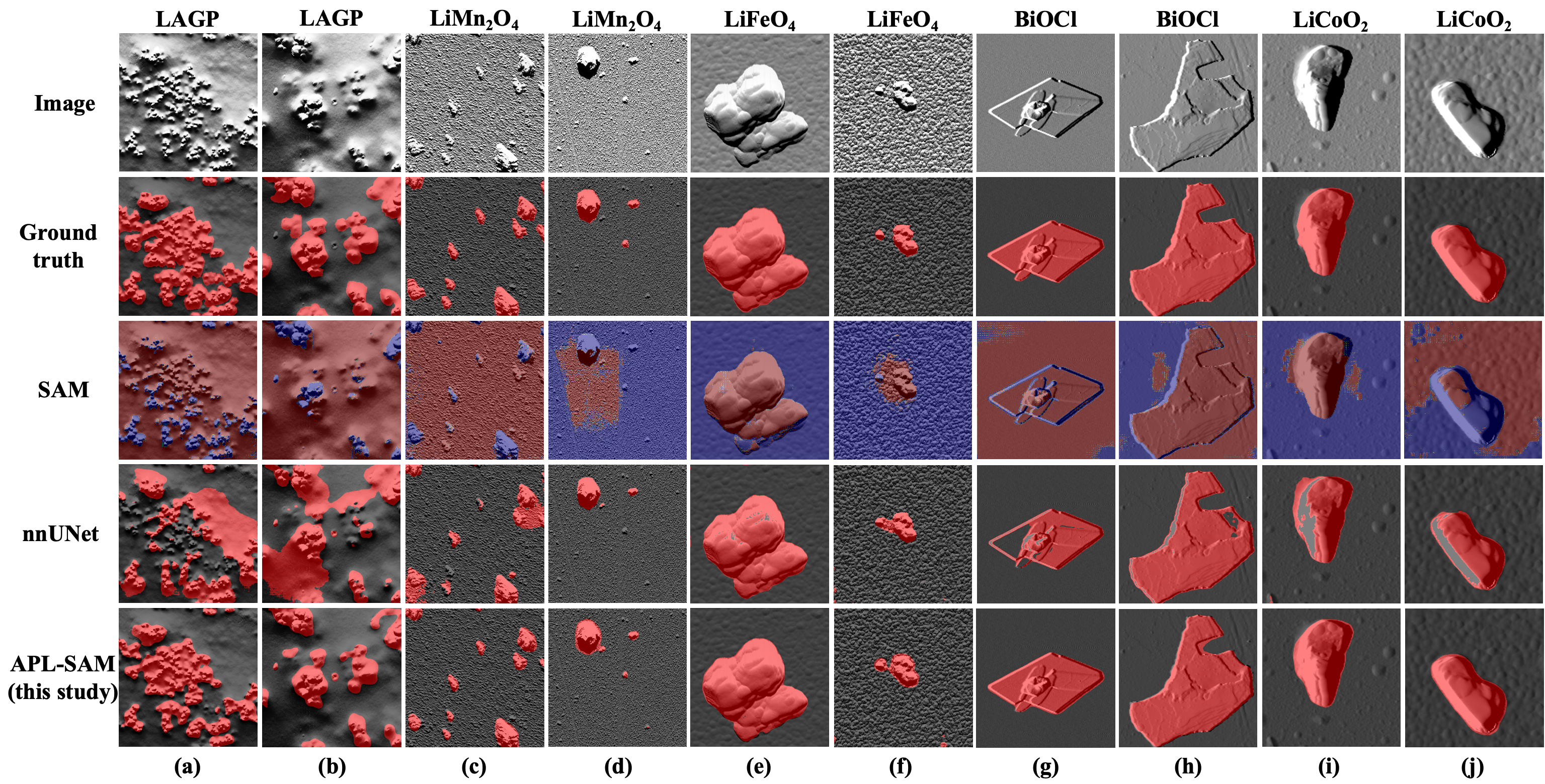}
\caption{Qualitative results of the segmentation methods in comparison. The $1^{st}$ row represents the raw input query images and the $2^{nd}$ row denotes the corresponding ground truth segmentations, where the target objects are visualized in red. The $3^{rd}$ row presents SAM's predictions with 20-point prompts, where foreground regions are marked in red and background regions in blue. The $4^{th}$ row depicts predictions from nnUNet, and the $5^{th}$ row shows predictions from the APL-SAM presented in this study, with target objects highlighted in red in both rows. Here, the scan size for sample $LAGP(a)$ is $20\mu m \times20\mu m$, and the scan size for samples $LiFeO_{4}(e)$ and $LiCoO_{2}(j)$ is $1\mu m \times1\mu m$. For all other samples, the scan size is $5\mu m \times5\mu m$.}
\label{fig_6}
\end{figure*}

Fig. 6 presents segmentation masks produced by SAM (with 20-point prompts), nnUNet, and the proposed APL-SSM. Notably, SAM struggles with segmenting smaller particles and distinguishing objects with unclear foreground-background differences, such as $LiMn_{2}O_{4}$ in column (d), $LiFeO_4$ in column (f) and $LiCoO_2$ in column (i). For samples with multiple target objects or uneven object distribution, SAM's performance tends to deteriorate further, as illustrated in column (a) to column (c). As for nnUNet, despite having seen all the training samples, it still performs sub-optimally in boundary regions and multi-particle scenarios. For example, the segmentation results for $LiMn_{2}O_{4}$ shown in the column (c) reveal that although nnUNet is capable of segmenting multiple targets, it still struggles with boundary accuracy and exhibits instances of missed detections. In columns (i) and (j), nnUNet is affected by shadows caused by differential signals, resulting in a noticeable number of false negatives. In contrast, APL-SAM accurately segment both single and multiple objects with improved precision at the boundary of objects and also robustness. Notably, APL-SAM also demonstrates exceptional performance on more challenging segmentation instances, such as those uneven surface and distribution involving $LAGP$ (Column (a) and (b)).
\subsection{Ablation Studies}
\subsubsection{Effectiveness of the Adaptive Prompt Learning strategy}
We first conduct ablation studies to evaluate the effectiveness of the Adaptive Prompt Learning strategy. To validate the effectiveness of the Adaptive Prompt Learning (APL) strategy, we compare its performance against several alternative prompt encoder configurations. First, we remove the prompt encoder entirely, relying solely on image embeddings and the support mask for segmentation (denoted as “No PE” in Table III). Second, we replace APL with a single holistic prototype generated via masked average pooling (“1 Prototype” in Table III). Third, we directly use point coordinates as prompt (“Point Prompt” in Table III), ensuring that the number of points is consistent with $N_{max}$ for a fair comparison. Table III presents the comparison of the performance from these four different strategies in terms of both DSC and IoU. It can be observed that prompt-based networks reliably surpass the architecture without a prompt encoder. The results demonstrate that our APL consistently outperforms other strategies across all five materials used in this study.
\begin{table*}[!t]
\caption{Performance Comparisons of different prompt encoder strategies in terms of DSC and IoU. The \textbf{best} and \underline{second best} results are highlighted with \textbf{bold} and \underline{underline}, respectively.\label{tab:table3}}
\centering
\begin{tabular}{|c|c|c|c|c|c|c|c|c|c|c|}
\hline
\multirow{2}{*}{Prompt Strategy} & \multicolumn{2}{c|}{$LiFeO_{4}$} & \multicolumn{2}{c|}{$LiMn_{2}O_{4}$} & \multicolumn{2}{c|}{$LAGP$} & \multicolumn{2}{c|}{$BiOCl$} & \multicolumn{2}{c|}{$LiCoO_{2}$}\\
\cline{2-11}
  & DSC & IoU & DSC & IoU & DSC & IoU & DSC & IoU & DSC & IoU\\
\hline
No PE & 91.67 & 84.63 & 91.32 & 84.03 & 71.34 & 55.44 & 88.00 & 78.56 & 92.01 & 85.21\\
1 Prototype & 92.75 & 86.49 & \underline{91.74} & \underline{84.74} & 72.25 & 56.55 & 88.73 & 78.62 & \underline{93.95} & \underline{88.59}\\
Point Prompt & \underline{94.00} & \underline{88.67} & 91.55 & 84.42 & \underline{72.84} & \underline{57.28} & \underline{89.11} & \underline{80.36} & 93.08 & 87.06\\
APL & \textbf{95.35} & \textbf{91.12} & \textbf{92.60} & \textbf{86.22} & \textbf{73.54} & \textbf{58.16} & \textbf{90.92} & \textbf{83.35} & \textbf{95.24} & \textbf{90.91}\\
\hline
\end{tabular}
\end{table*}

\subsubsection{Effectiveness of the Multi-Level Mask Decoder}
Additionally, the impact of the Multi-Level Mask Decoder (MLMD) is also thoroughly assessed. Tables I and II also includes the performance of APL-SAM after removing multi-level mask decoder (MLMD), where "without MLMD" means disabling the parallel operation of multiple attention-based blocks. It can be seen that, without MLMD, the average DSC across the five classes decreases by 1.64\%, and the average IoU drops by 2.95\%, further validating the effectiveness of MLMD used in this study.
\begin{figure}[!t]
\centering
\includegraphics[width=3.4in]{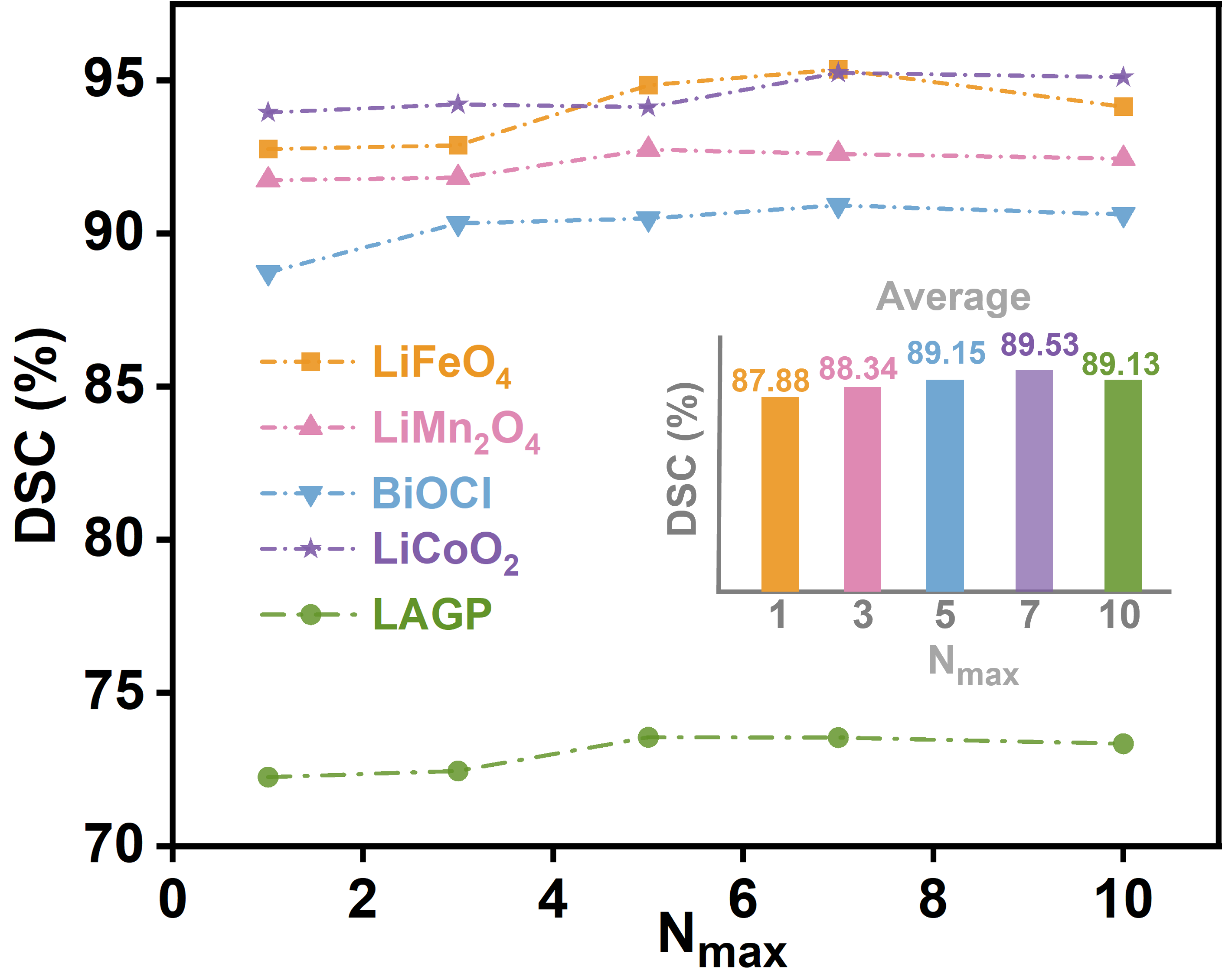}
\caption{Hyperparameter experiments on the maximum number of the centroids $N_{max}$.}
\label{fig_7}
\end{figure}

\subsubsection{Effectiveness of the maximal number of visual prompts $N_{max}$}
To manage computational resources, this study imposes a maximum limit on the number of visual prompts (equivalent to superpixel centroids) in the Adaptive Prompt Learning module. For target objects that are not exceptionally small, the APL module selects the maximum feasible number of centroids. The ablation studies on this hyperparameter, $N_{max}$, is presented in Fig. 7. It shows clearly that the DSC results for various $N_{max}$ values for different materials. The highest DSC for $LiMn_{2}O_{4}$ is achieved with $N_{max}=5$, while the optimal results for the other four classes are obtained with $N_{max}=7$. Averaging results across all classes, $N_{max}=7$ consistently yields the best performance. Consequently, we set $N_{max}=7$ for our model.

\section{Conclusion}
In this study, to address the suboptimal performance of SAM in SPM image segmentation, we propose an enhanced framework based on few-shot learning, named APL-SAM, enabling SAM to achieve high-precision segmentation of unseen classes in SPM dataset under one-shot guidance. Our enhanced framework consists of three key components: 1) incorporating lightweight Adapters into SAM's image encoder to better learn and adapt to the features of the SPM images; 2) using an Adaptive Prompt Learning-based prompt encoder to generate superpixel centroids as visual prompts through a superpixel-based clustering method; and 3) upgrading SAM’s mask decoder to a multi-level mask decoder which integrates multi-layer query embeddings and support masks to yield comprehensive segmentation results. Experimental results demonstrate that APL-SAM effectively achieve precise semantic segmentation in SPM images, providing strong validation for the model's efficacy as proposed in this study. Additionally, we introduce the SPM-Seg dataset for SPM image segmentation, which includes key materials for battery systems and it can be served as a valuable resource for advancing the deep-learning-based SPM image processing tasks.

\bibliographystyle{IEEEtran}
\bibliography{manuscript}


%

\vfill

\end{document}